\documentclass{article}

\usepackage[final]{neurips_2024}
\usepackage[utf8]{inputenc} 
\usepackage[T1]{fontenc}    
\usepackage{hyperref}       
\usepackage{url}            
\usepackage{booktabs}       
\usepackage{amsfonts}       
\usepackage{nicefrac}       
\usepackage{microtype}      
\usepackage{amsmath}        
\usepackage{xcolor}         
\usepackage{graphicx}       
\usepackage{listings}       
\usepackage{subcaption}     
 
\colorlet{punct}{red!60!black}
\definecolor{background}{HTML}{EEEEEE}
\definecolor{delim}{RGB}{20,105,176}
\colorlet{numb}{magenta!60!black}

\lstdefinelanguage{json}{
    basicstyle=\normalfont\ttfamily\footnotesize,
    numbers=left,
    numberstyle=\scriptsize,
    stepnumber=1,
    numbersep=8pt,
    showstringspaces=false,
    breaklines=true,
    frame=lines,
    backgroundcolor=\color{background},
    literate=
     *{0}{{{\color{numb}0}}}{1}
      {1}{{{\color{numb}1}}}{1}
      {2}{{{\color{numb}2}}}{1}
      {3}{{{\color{numb}3}}}{1}
      {4}{{{\color{numb}4}}}{1}
      {5}{{{\color{numb}5}}}{1}
      {6}{{{\color{numb}6}}}{1}
      {7}{{{\color{numb}7}}}{1}
      {8}{{{\color{numb}8}}}{1}
      {9}{{{\color{numb}9}}}{1}
      {:}{{{\color{punct}{:}}}}{1}
      {,}{{{\color{punct}{,}}}}{1}
      {\{}{{{\color{delim}{\{}}}}{1}
      {\}}{{{\color{delim}{\}}}}}{1}
      {[}{{{\color{delim}{[}}}}{1}
      {]}{{{\color{delim}{]}}}}{1},
}

\title{Semantic-Drive: Democratizing Long-Tail Data Curation via Open-Vocabulary Grounding and Neuro-Symbolic VLM Consensus}

\author{%
  Antonio Guillen-Perez  \\
  Independent Researcher \\
  \texttt{antonio\_algaida@hotmail.com} \\
  \href{https://antonioalgaida.github.io/}{antonioalgaida.github.io}
}

\begin{document}

\maketitle
\begin{abstract}
The development of robust Autonomous Vehicles (AVs) is bottlenecked by the scarcity of "Long-Tail" training data. While fleets collect petabytes of video logs, identifying rare safety-critical events (e.g., erratic jaywalking, construction diversions) remains a manual, cost-prohibitive process. Existing solutions rely on coarse metadata search, which lacks precision, or cloud-based VLMs, which are privacy-invasive and expensive. We introduce \textbf{Semantic-Drive}, a local-first, neuro-symbolic framework for semantic data mining. Our approach decouples perception into two stages: (1) \textbf{Symbolic Grounding} via a real-time open-vocabulary detector (YOLOE) to anchor attention, and (2) \textbf{Cognitive Analysis} via a Reasoning VLM that performs forensic scene analysis. To mitigate hallucination, we implement a "System 2" inference-time alignment strategy, utilizing a multi-model "Judge-Scout" consensus mechanism. Benchmarked on the nuScenes dataset against the Waymo Open Dataset (WOD-E2E) taxonomy, Semantic-Drive achieves a \textbf{Recall of 0.966} (vs. 0.475 for CLIP) and reduces Risk Assessment Error by \textbf{40\%} compared to the best single scout models. The system runs entirely on consumer hardware (NVIDIA RTX 3090), offering a privacy-preserving alternative to the cloud.
\end{abstract}

\section{Introduction}

\begin{figure}[t]
    \centering
    \includegraphics[width=0.6\textwidth]{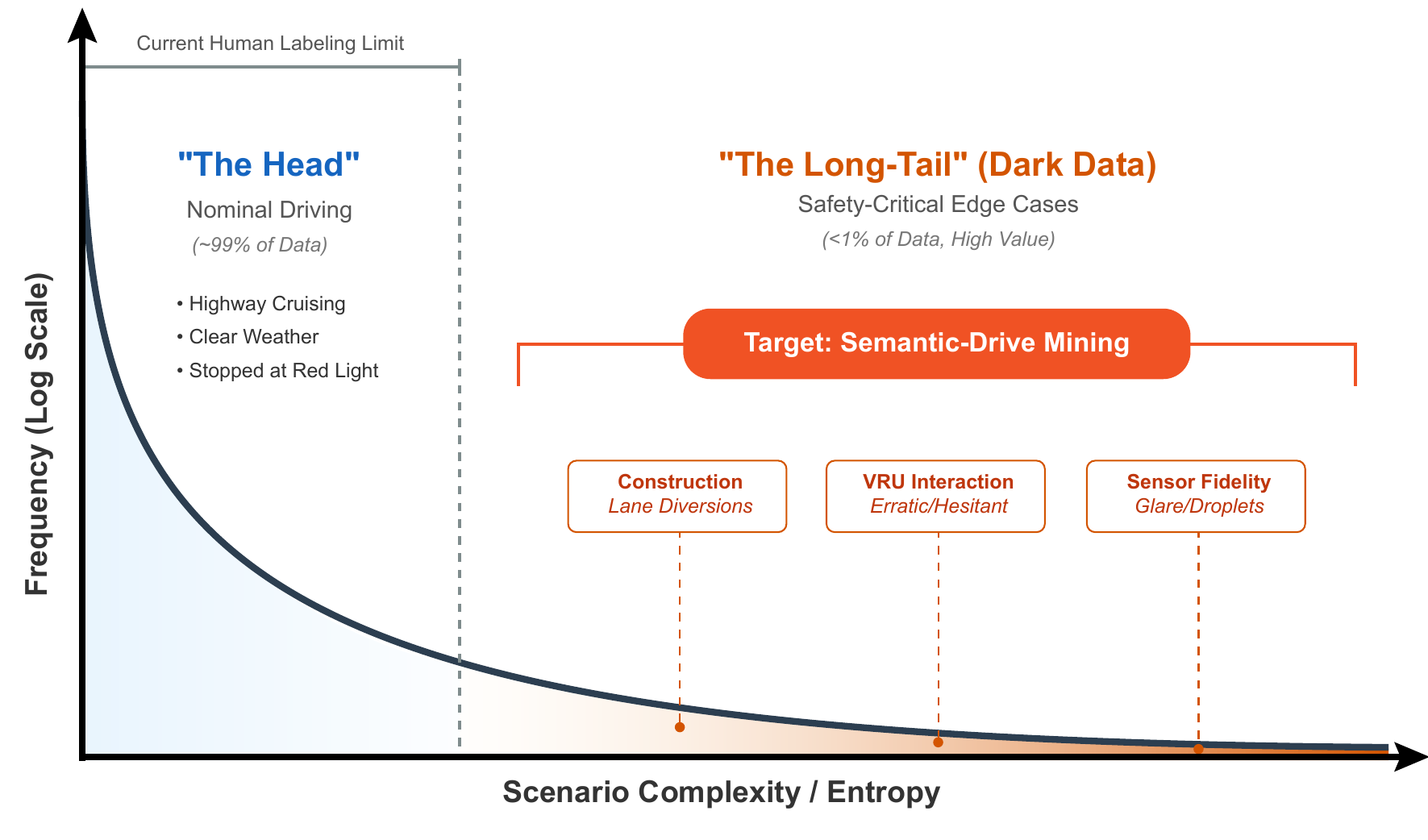}
    \caption{\textbf{The "Dark Data" Crisis in Autonomous Driving.} 
    The distribution of driving scenarios follows a Power Law (Zipfian) distribution. 
    \textbf{(Left) The "Head":} Represents 99\% of data logs, consisting of nominal, low-entropy driving (e.g., highway cruising) which provides diminishing returns for model training.
    \textbf{(Right) The "Long Tail":} Contains rare, safety-critical edge cases defined by the Waymo Open Dataset (WOD-E2E) taxonomy, such as erratic VRUs or sensor degradation. 
    Traditional human annotation is cost-prohibitive for mining this region. \textbf{Semantic-Drive} automates the retrieval of these high-value samples.}
    \label{fig:long_tail}
\end{figure}

The fundamental challenge in scaling autonomous perception is the \textit{imbalanced distribution} of training data. As illustrated in Figure \ref{fig:long_tail}, driving scenarios follow a heavy-tailed (Zipfian) distribution. The "Head" of the distribution comprises the vast majority of collected logs ($\approx$99\%), representing nominal driving conditions such as highway cruising or stopped traffic. While abundant, this data offers diminishing returns for improving model robustness.

The critical value for Level 4 safety validation lies in the "Long Tail" rare, high-entropy events such as construction zones with conflicting lane markings, erratic vulnerable road users (VRUs), or sensor degradation due to sudden weather changes \cite{caesar2020nuscenes}. Currently, identifying these samples within petabyte-scale "Data Lakes" constitutes a "Dark Data" crisis. Manual review is cost-prohibitive at this scale, and heuristic metadata tags (e.g., \texttt{weather=rain}) lack the semantic granularity to distinguish between a wet road and a dangerous hydroplaning risk.

Currently, mining these safety-critical scenarios from archival footage is a bottleneck. Traditional methods rely on brittle heuristics (e.g., querying CAN bus data for hard braking) or metadata keyword search, which suffers from poor temporal granularity. While recent Vision-Language Models (VLMs) like GPT-4V offer promising semantic understanding, relying on closed-source cloud APIs for data curation is impractical for the automotive industry due to strict data privacy regulations (GDPR), bandwidth constraints, and the prohibitive cost of processing video streams at scale.

To bridge this gap, we introduce \textbf{Semantic-Drive}, a privacy-preserving, local-first framework for semantic data mining. Unlike end-to-end driving agents (e.g., DriveGPT4 \cite{xu2024drivegpt4}) that utilize VLMs for \textit{control}, Semantic-Drive focuses on \textit{retrieval}, acting as a "Cognitive Indexer" that transforms raw, unstructured video logs into a queryable semantic database.

\textbf{Semantic-Drive} is a novel \textbf{Neuro-Symbolic Architecture} designed to run efficiently on consumer-grade hardware (e.g., a single NVIDIA RTX 3090). Pure VLMs often suffer from hallucination and "small object blindness." To mitigate this, our framework separates perception into two distinct pathways: (1) A symbolic "Grounding" stage using real-time Open-Vocabulary Object Detection to generate a high-recall inventory of hazards, and (2) A cognitive "Reasoning" stage where a Chain-of-Thought (CoT) VLM performs forensic analysis to verify detections and assess risk.

Our contributions are as follows:
\begin{itemize}
    \item \textbf{Neuro-Symbolic "System 2" Architecture:} We introduce an inference-time alignment strategy inspired by System 2 reasoning \cite{bengio2019system2}. By enforcing a "Skepticism Policy" (where the VLM must logically verify symbolic detections against visual evidence) we significantly reduce hallucinations compared to standard zero-shot prompting, \textbf{lowering the Risk Assessment Error (MAE) by 51\%}, compared to the baseline Pure VLM.
    \item \textbf{Judge-Scout Consensus Mechanism:} We address the stochastic nature of LLMs through a multi-model consensus engine. We demonstrate that aggregating reasoning traces from heterogeneous scouts (e.g., Qwen3-VL, Gemma3, Kimi-VL) via a "Judge" reduces Risk Assessment Error (MAE) from 1.13 to 0.676.
    \item \textbf{Mapping the WOD-E2E Taxonomy:} We convert the Waymo Open Dataset (WOD-E2E) taxonomy \cite{wod_e2e} into a structured JSON schema. This allows automated extraction of detailed causal attributes, such as "Implicit Lane Diversion" and "Sensor Fidelity Issues," which standard CLIP embeddings often miss.  
    \item \textbf{Efficient, Local Data Curation:} We show that high-quality scenario curation can be done entirely on consumer hardware. Our local pipeline cuts the marginal cost of curation by approximately 97\% compared to cloud-based solutions, making advanced data mining accessible without relying on external services.
\end{itemize}

The complete source code and evaluation scripts are available in our GitHub repository\footnote{\url{https://github.com/AntonioAlgaida/Semantic-Drive}}. To facilitate reproducibility, we release the full generated semantic index ($N=2,550$) and the annotated Gold Set via Hugging Face Datasets\footnote{\url{https://huggingface.co/datasets/agnprz/semantic-drive-results}}. Additionally, we host an interactive 'Data Explorer' visualizing the mined scenarios on Hugging Face Spaces\footnote{\url{https://huggingface.co/spaces/agnprz/Semantic-Drive-Explorer}}.

\section{Related Work}

\subsection{Vision-Language Models in Autonomous Driving}
The integration of Large Multimodal Models (LMMs) into the autonomous driving stack has largely bifurcated into two streams: \textit{generative simulation} and \textit{end-to-end control}. Systems such as DriveGPT4 \cite{xu2024drivegpt4}, DriveVLM \cite{tian2024drivevlm}, and Waymo's EMMA \cite{waymo_emma} leverage VLMs to map sensor inputs directly to control actions (e.g., steering angle) or to generate natural language explanations for driving decisions. 

However, these "VLM-as-Agent" approaches utilize reasoning for \textit{execution}, not \textit{curation}. They are computationally intensive and require massive, pre-curated instruction datasets for fine-tuning. They are ill-suited for the upstream task of filtering petabytes of raw logs due to their high inference latency and cost. \textbf{Semantic-Drive} complements this ecosystem by acting as the foundational \textit{DataOps} engine, automating the discovery of the high-value, long-tail training samples required to robustify these downstream driving agents.

\subsection{Semantic Data Mining \& Scenario Retrieval}
Retrieving safety-critical edge cases from unlabelled "Dark Data" remains an open challenge. Current methodologies fall into three categories:

\begin{itemize}
    \item \textbf{Heuristic Metadata Search:} Standard datasets like nuScenes \cite{caesar2020nuscenes} rely on manual tags or CAN-bus triggers (e.g., hard braking). As shown in our experiments, this suffers from poor temporal granularity, flagging entire scenes based on transient events.
    \item \textbf{Programmatic Mining:} Approaches like RefAV \cite{davidson2025refav} synthesize programmatic queries (e.g., SQL or Python) to filter track-level data. While efficient, they are "Semantically Blind", relying on geometric primitives (bounding boxes) while missing visual nuances such as construction signage, debris types, or pedestrian gaze direction.
    \item \textbf{Latent Embedding Search:} Methods like VLMine \cite{ye2025vlmine} and localized CLIP searches utilize vector similarity to find scenarios. However, global embeddings suffer from "Bag-of-Words" blindness; they often fail to distinguish between a pedestrian \textit{on the sidewalk} (safe) and one \textit{in the lane} (hazard) due to a lack of spatial binding \cite{Zhong2021Dec}.
\end{itemize}

In contrast, \textbf{Semantic-Drive} introduces a \textit{Causal Reasoning Layer}. Instead of relying on geometric tracks or statistical keyword frequency, we employ Chain-of-Thought (CoT) reasoning to analyze the \textit{implications} of a scene (e.g., "The orange drums are forcing a lane merge"), enabling the retrieval of scenarios defined by complex causal interactions.

\begin{table}[h]
  \centering
  \caption{Qualitative Comparison with State-of-the-Art Mining Approaches. Semantic-Drive is distinct in its ability to perform privacy-preserving, pixel-level causal reasoning without reliance on pre-computed tracks or cloud APIs.}
  \label{tab:baselines}
  \resizebox{\textwidth}{!}{%
  \begin{tabular}{lccccc}
    \toprule
    \textbf{Method} & \textbf{Input Modality} & \textbf{Privacy (Local)} & \textbf{Reasoning Level} & \textbf{Requires Tracks?} & \textbf{Spatial Logic?} \\
    \midrule
    RefAV \cite{davidson2025refav} & Metadata/Tracks & \checkmark & Geometric (Speed/Pos) & Yes & \checkmark \\
    VLMine \cite{ye2025vlmine} & Images (Cloud) & No & Statistical (Frequency) & No & No \\
    CLIP Embeddings & Images (Local) & \checkmark & Semantic Similarity & No & No \\
    \textbf{Semantic-Drive (Ours)} & \textbf{Raw Pixels} & \textbf{\checkmark} & \textbf{Causal/Forensic} & \textbf{No} & \textbf{Yes} \\
    \bottomrule
  \end{tabular}%
  }
\end{table}

\subsection{Neuro-Symbolic Grounding \& Hallucination Mitigation}
A major barrier to deploying VLMs in safety-critical domains is hallucination, specifically, the tendency to invent objects or misinterpret spatial relationships. To mitigate this, recent computer vision research has adopted neuro-symbolic architectures that pair a "Symbolic" detector with a "Cognitive" reasoner. Open-vocabulary detectors like GLIP \cite{li2022grounded} and YOLO-World \cite{cheng2024yolow} allow for the grounding of arbitrary text prompts.

\textbf{Semantic-Drive} builds upon this by employing a "Prompt Injection" strategy. We utilize the output of a high-recall open-vocabulary segmentor (YOLOE \cite{wang2025yoloerealtimeseeing}) not as the final result, but as a \textit{grounded attention anchor} for the VLM. Unlike zPROD \cite{zprod2025}, which uses VLMs primarily for bounding box refinement, our framework enforces a "System 2" \textit{Skepticism Policy}: the VLM is tasked with logically verifying the symbolic inventory against the visual evidence, using the detector's confidence scores to weigh conflicting signals. This bidirectional validation significantly reduces false positives compared to pure neural approaches.

\section{Methodology}

The Semantic-Drive framework is designed as a local-first, privacy-preserving DataOps engine. Unlike cloud-based solutions that process video streams via API, our architecture operates entirely on consumer-grade hardware, specifically a single NVIDIA RTX 3090 with 24GB VRAM. The pipeline transforms raw, synchronized multi-camera feeds into a structured semantic database through a three-stage Neuro-Symbolic process: (1) Symbolic Grounding, (2) Cognitive Analysis, and (3) Multi-Model Consensus.

\begin{figure*}[t]
    \centering
    \includegraphics[width=\textwidth]{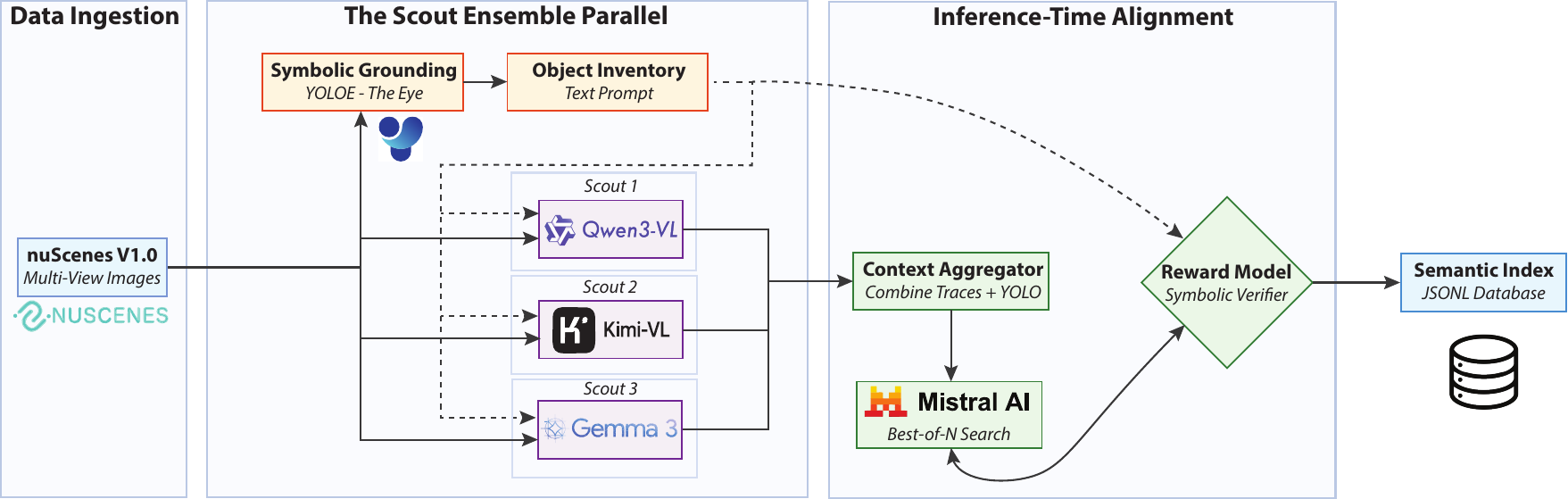}
    \caption{\textbf{The Semantic-Drive System Architecture.} 
    The pipeline operates in three stages: 
    (1) \textbf{Symbolic Grounding (Orange):} YOLOE extracts a textual object inventory from raw frames. 
    (2) \textbf{Cognitive Scouting (Purple):} An ensemble of heterogeneous Reasoning VLMs (Qwen, Kimi, Gemma) perform independent forensic analysis, anchored by the symbolic inventory. 
    (3) \textbf{Consensus \& Alignment (Green):} The Judge (Ministral-3-14B) synthesizes the scout reports. A deterministic Reward Model performs inference-time verification (\textit{Best-of-N}) to filter hallucinations before committing to the final Semantic Index.}
    \label{fig:architecture}
\end{figure*}

\subsection{Stage 1: Symbolic Grounding (The Eye)}
To mitigate the small object blindness and spatial hallucinations common in pure Vision-Language Models (VLMs), we employ a strong symbolic prior. We utilize \textbf{YOLOE} \cite{wang2025yoloerealtimeseeing}, a real-time open-vocabulary segmentation model, to perform an initial sweep of the visual field.

\textbf{WOD-E2E Taxonomy Alignment.} Instead of generic COCO classes, we inject a custom text-prompt taxonomy aligned with the Waymo Open Dataset for End-to-End Driving (WOD-E2E) \cite{wod_e2e}. This includes specific long-tail categories such as \texttt{orange drum}, \texttt{jersey barrier}, \texttt{debris}, \texttt{puddle}, and \texttt{construction worker}. More information about the extracted classes can be found in \ref{yolo_voc}.

\textbf{The Object Inventory.} The detector output is not saved directly. Instead, it is converted into a structured textual \textit{Object Inventory} that serves as a prompt injection for the VLM. Formally, we define the inventory set $\mathcal{I}$ as:

\begin{equation}
     \mathcal{I} = \{ (Class_i, Cam_j, c_i, S_{rel}) \mid c_i > \tau_{recall} \}
\end{equation}

Where:
\begin{itemize}
    \item $Class_i$: The semantic category mapped to the WOD-E2E taxonomy.
    \item $Cam_j$: The spatial origin of the detection (Front-Left, Center, or Front-Right).
    \item $c_i$: The model confidence score.
    \item $S_{rel}$: The Relative Size, calculated as the ratio of bounding box area to total image resolution ($S_{rel} = \frac{Area_{bbox}}{H \times W}$). This serves as a heuristic proxy for proximity depth, where objects with $S_{rel} > 0.1$ are encoded as "Large" to signal immediate proximity.
\end{itemize}

\begin{figure*}[t] 
    \centering
    \includegraphics[width=\textwidth]{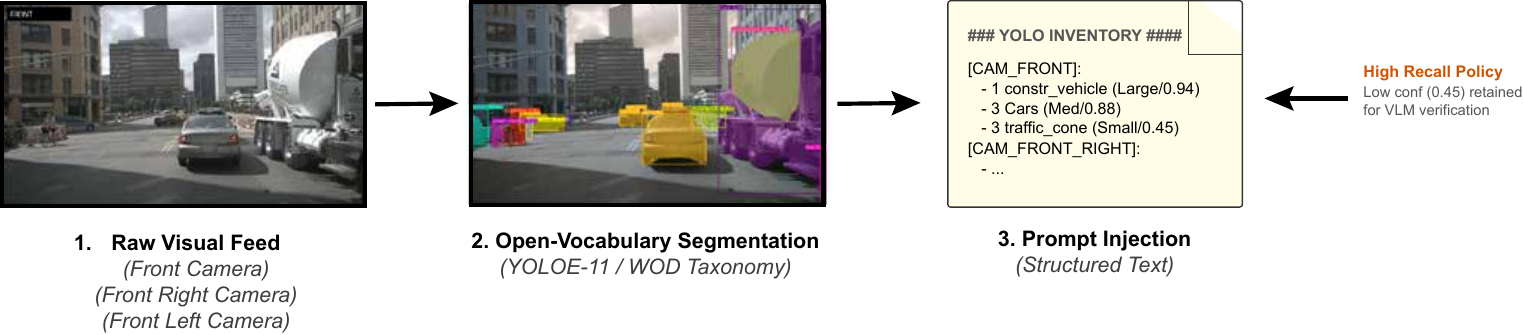}
    \caption{\textbf{Stage 1: Symbolic Grounding (Real-World Inference).} 
    From Pixels to Prompts. 
    \textbf{(Left)} The raw Front-Center camera feed showing a complex construction scenario. 
    \textbf{(Center)} The segmentation mask identifies the cement mixer (Purple) and a distant traffic cone (Orange). Note the low confidence score ($0.45$) on the cone.
    \textbf{(Right)} Instead of filtering this "weak" detection, our High-Recall Policy retains it in the textual inventory (highlighted orange). This allows the downstream VLM to act as the final arbiter, preventing the "False Negative" blindness common in traditional rigid thresholds.}
    \label{fig:grounding}
\end{figure*}

We empirically calibrated a low confidence threshold ($\tau_{recall} = 0.15$) to prioritize Recall over Precision at the symbolic stage. This design deliberately admits potential false positives, such as reflections or billboards, into the context window to ensure that subtle long-tail objects are not filtered out early. The burden of False Positive Rejection is thus delegated to the downstream reasoning engine which utilizes contextual skepticism to filter these artifacts. An example of the Stage 1 pipeline can be found on \ref{fig:grounding}.

\subsection{Stage 2: Cognitive Analysis (The Brain)}
The core of our framework is the reasoning engine where a quantized VLM (e.g., Qwen3-VL-30B or Kimi-Thinking) performs forensic analysis. We employ a \textbf{Front-Hemisphere Attention Strategy} by processing synchronized Front-Left, Front-Center, and Front-Right images to maximize resolution within the context window while discarding less relevant rear views.

\textbf{Neuro-Symbolic Prompt Injection.} The VLM receives both the raw pixel data and the \textit{Object Inventory}. We enforce a \textbf{"Skepticism Policy"} via the system prompt to handle the low-confidence detections from Stage 1:
\begin{itemize}
    \item \textbf{High Confidence ($c > 0.8$):} The VLM is instructed to trust the detection but verify the semantic context, such as verifying if a pedestrian is interacting with the scene.
    \item \textbf{Low Confidence ($c < 0.5$):} The VLM treats the detection as a hypothesis. It must perform a visual verification step to reject artifacts, for example, noting that YOLO sees a person but visual evidence shows a static poster.
\end{itemize}

\textbf{Scenario DNA Schema.} Unlike simple tagging, we extract a hierarchical JSON structure describing the causal physics of the scene. The schema covers four layers:
\begin{enumerate}
    \item \textbf{ODD Attributes:} Weather, Lighting, and Sensor Fidelity.
    \item \textbf{Topology:} Map divergence indicators like physical lane restrictions.
    \item \textbf{Actor Dynamics:} Behavioral states such as hesitation rather than simple presence.
    \item \textbf{Planner Logic:} The implied ego-maneuver, including nudging or emergency braking.
\end{enumerate}

\begin{figure*}[t]
    \centering
    \includegraphics[width=0.9\textwidth]{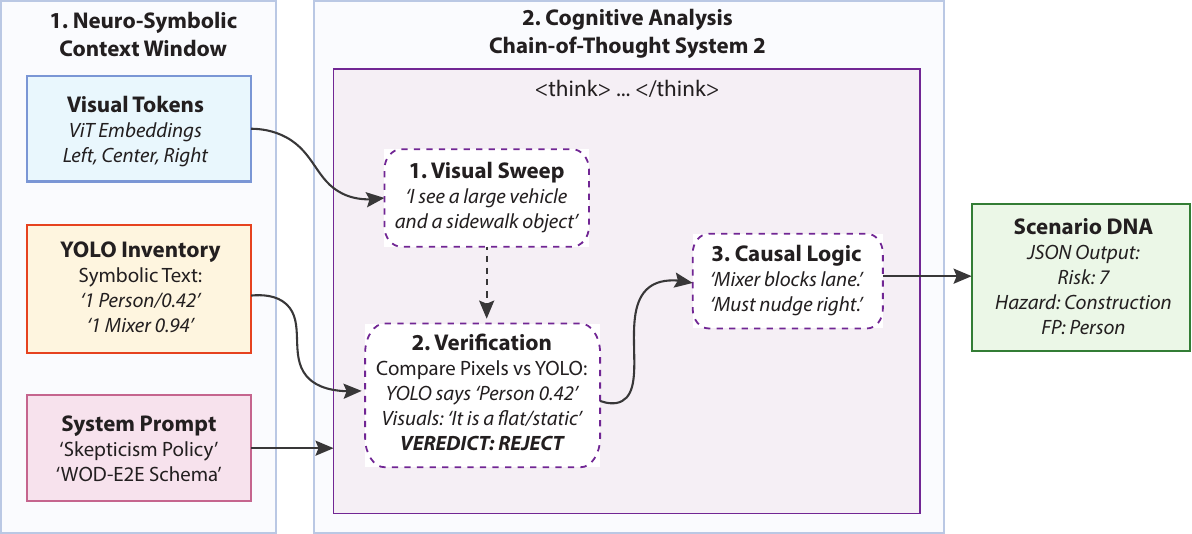}
    \caption{\textbf{Stage 2: The Cognitive Reasoning Engine.} 
    A visualization of the single-scout inference process. 
    \textbf{(Left)} The model ingests a hybrid context of Visual Tokens (Blue), Symbolic Text (Orange), and the System Prompt (Pink).
    \textbf{(Center)} Instead of immediate generation, the model enters a "System 2" thinking phase (Chain-of-Thought). Here, it executes the \textit{Skepticism Policy}: explicitly comparing the low-confidence YOLO detection against visual evidence to filter hallucinations (e.g., rejecting a poster classified as a person).
    \textbf{(Right)} Only after logical verification does it synthesize the final structured Scenario DNA.}
    \label{fig:cognitive_process}
\end{figure*}

The complete prompt can be found in \ref{whole_prompt}.
The process is detailed in the Figure \ref{fig:cognitive_process}

\subsection{Stage 3: Multi-Model Consensus (The Judge)}
Single VLMs are stochastic and prone to bias. To ensure high-fidelity curation without human-in-the-loop, we introduce a \textbf{Judge-Scout Architecture}.

We deploy multiple heterogeneous "Scout" models (e.g., Qwen3-VL and Kimi-Thinking) to process the same frame in parallel. A separate "Judge" LLM (Ministral-3-14B) aggregates their JSON outputs and Reasoning Traces. The Judge applies a \textbf{Safety-Bias Voting Logic}:
\begin{itemize}
    \item If Scouts disagree on a high-risk attribute, the Judge favors the positive detection provided valid reasoning is present in the trace.
    \item If a Scout proposes a rare object that is uncorroborated by the YOLO inventory or the peer Scout, it is flagged for rejection or manual review.
\end{itemize}
This mechanism significantly reduces false negatives for safety-critical events while filtering out idiosyncratic hallucinations. The exact system prompt enforcing these rules is detailed in \textbf{Appendix \ref{judge_prompt}}.

\subsection{Stage 4: Inference-Time Alignment (Symbolic Verification)}
Standard Large Language Models are stochastic, meaning a single generation may fail to adhere to strict schema constraints or hallucinate objects not present in the symbolic inventory. To mitigate this without expensive fine-tuning, we implement an inference-time search strategy inspired by \textit{Best-of-N} sampling.

The Consensus Judge generates $N=3$ candidate scenario descriptions. We rank these candidates using a deterministic, rule-based reward function $R(y, \mathcal{I}_{yolo})$ that penalizes objects not corroborated by the symbolic evidence provided by the high-recall detector:

\begin{equation}
    R(y) = \alpha \cdot \mathbb{I}_{grounding} + \beta \cdot \mathbb{I}_{causality} - \gamma \cdot \mathbb{I}_{hallucination}
\end{equation}

We empirically set the hyperparameters to prioritize safety over recall: \textbf{$\alpha=2.0$} (Grounding Reward), \textbf{$\beta=3.0$} (Causal Consistency Reward), and \textbf{$\gamma=10.0$} (Hallucination Penalty). The high penalty for hallucinations ($\gamma$) ensures that the model aggressively discards candidates that invent critical hazards unsupported by the symbolic inventory.

Where:
\begin{itemize}
    \item $\mathbb{I}_{grounding}$: Reward if critical tags are corroborated by the YOLO Inventory.
    \item $\mathbb{I}_{causality}$: Reward if a planner action is linked to a valid blocking factor.
    \item $\mathbb{I}_{hallucination}$: Heavy penalty if the VLM invents objects that appear in neither the YOLO inventory nor the Scout reports.
\end{itemize}

The system outputs the candidate $y^* = \text{argmax}_{y} R(y)$. This symbolic verification loop effectively acts as a "Unit Test" for the generated data to ensure the final database entry is logically sound.

\textbf{Connection to System-2 Reasoning.} While recent Large Reasoning Models like OpenAI o1 utilize implicit, learned reward models to guide internal chains of thought, these are often opaque and non-deterministic. In contrast, Semantic-Drive employs \textbf{Neuro-Symbolic Test-Time Compute}. Drawing on the cognitive distinction between intuitive (System 1) and deliberate (System 2) thinking \cite{kahneman2011thinking, bengio2019system2}, we utilize an \textbf{Explicit Outcome Reward Model (ORM)} implemented as a deterministic symbolic verifier. We generate $N$ candidate reasoning traces and perform \textbf{Rejection Sampling} based on consistency with the YOLO object inventory. This ensures that the deliberate "System 2" reasoning process is grounded in verifiable sensor data, strictly enforcing safety constraints that a purely neural reward model might overlook.

This implementation of 'Best-of-N' symbolic search is theoretically grounded in the findings of Walder et al. \cite{walder2025passk}, who demonstrated that for complex reasoning tasks with sparse solution spaces, optimizing for the joint utility of a set of samples ($Pass@k$) significantly outperforms standard optimization ($Pass@1$). While originally applied to Reinforcement Learning (RL) training, we adapt the principle to inference-time compute. By generating $N=3$ candidate reasoning traces and selecting the $\text{argmax}_{y} R(y)$ based on symbolic consistency, we effectively transform the VLM from a stochastic generator into a high-reliability $Pass@N$ estimator.

\begin{figure*}[t]
    \centering
    \includegraphics[width=\textwidth]{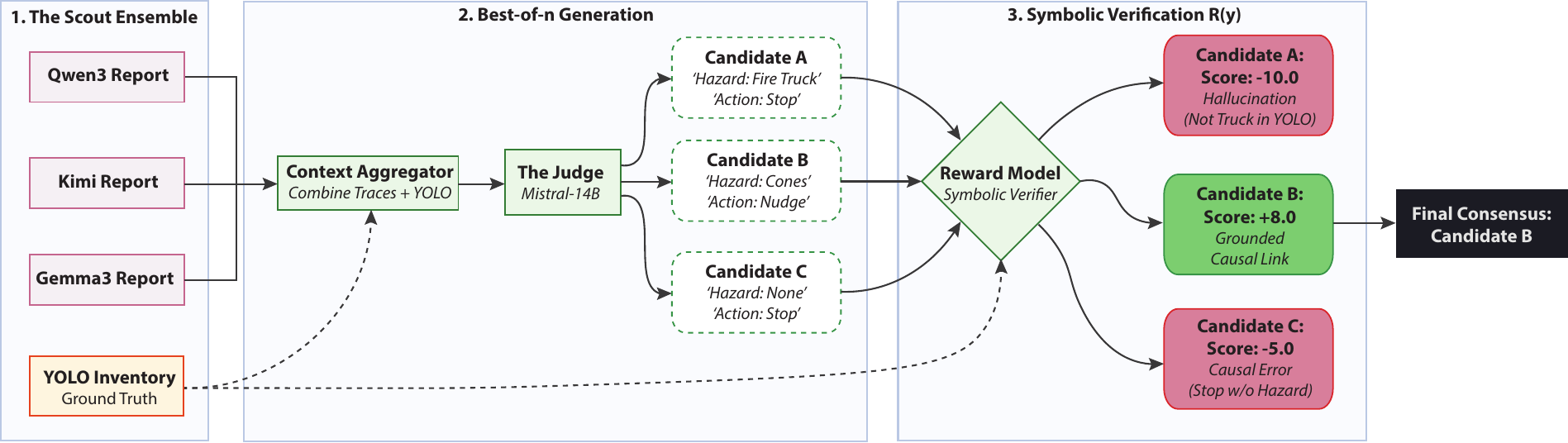}
    \caption{\textbf{Stage 3 \& 4: Inference-Time Alignment (The Judge).} 
    A visualization of the "System 2" consensus mechanism. 
    \textbf{(1)} The Judge (Ministral-3-14B) receives conflicting reports from the scouts. 
    \textbf{(2)} Instead of a single output, it performs a \textbf{Best-of-N Search}, generating multiple candidate scenarios. 
    \textbf{(3)} A deterministic \textbf{Symbolic Verifier} (Outcome Reward Model) scores each candidate against the YOLO inventory and causal logic constraints. In this example, Candidate A is rejected for hallucinating a "Fire Truck" not present in YOLO, while Candidate B is selected as the most grounded and logically consistent interpretation.}
    \label{fig:judge_consensus}
\end{figure*}

A detailed visual explanation of the judge consensus process and the Best-of-N process can be seen in Figure \ref{fig:judge_consensus}.

\subsection{Stage 5: Implementation Strategy}
To validate the feasibility of this architecture on consumer hardware, we implement the pipeline as a modular micro-service architecture:

\textbf{Inference Engine (The Scouts):} We utilize \texttt{llama.cpp} to serve quantized GGUF models. This allows 30B+ parameter Vision-Language Models to fit within the 24GB VRAM envelope of an RTX 3090 by utilizing 4-bit quantization (Q4\_K\_M), which retains 99\% of the reasoning performance of FP16 at 25\% of the memory cost.

\textbf{Consensus Node (The Judge):} The Judge operates asynchronously by processing the textual JSONL logs generated by the Scouts. To evaluate the trade-off between privacy and reasoning capability, we implement two configurations for the Consensus Node:
\begin{enumerate}
    \item \textbf{Local Judge:} An open-weights text-only LLM running locally. This ensures a 100\% air-gapped pipeline suitable for proprietary IP.
    \item \textbf{Cloud Oracle:} \textbf{Google Gemini 3 Pro}, used as a baseline to benchmark the quality of the local consensus.
\end{enumerate}
This dual-setup allows us to quantify the "Privacy Tax", or the performance gap, between fully local and cloud-assisted data curation.

\section{The "Scenario DNA" Taxonomy}

\begin{figure*}[t]
    \centering
    \includegraphics[width=0.85\textwidth]{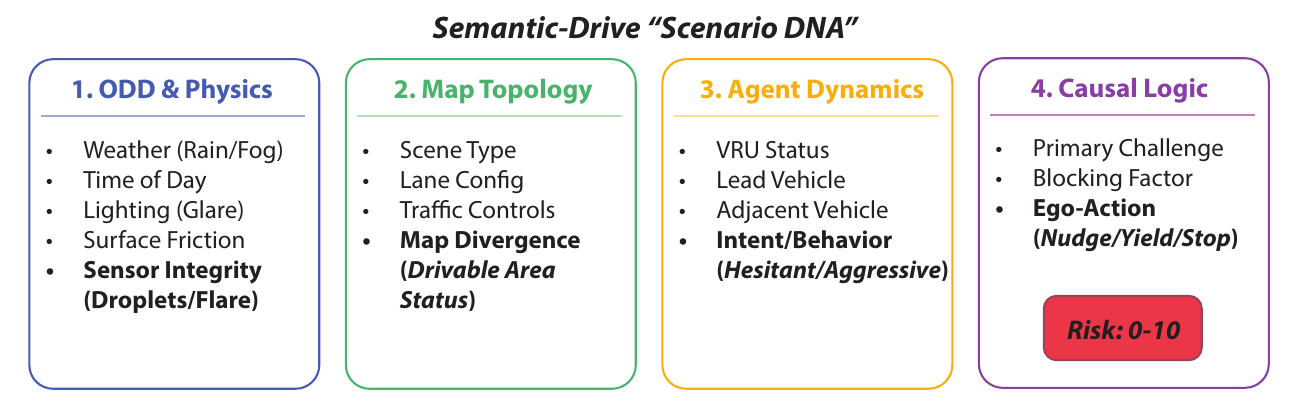} 
    \caption{\textbf{The "Scenario DNA" Hierarchical Taxonomy.} 
    Unlike standard flat metadata tags, Semantic-Drive enforces a causal dependency chain. 
    \textbf{(1) ODD \& Physics:} Defines environmental constraints, including novel attributes like \textit{Sensor Integrity} (e.g., lens flare).
    \textbf{(2) Map Topology:} Identifies contradictions between the HD Map and reality (\textit{Map Divergence}).
    \textbf{(3) Agent Dynamics:} Infers behavioral intent (e.g., hesitation) rather than just presence.
    \textbf{(4) Causal Logic:} Synthesizes the previous layers into a planner-centric risk assessment.}
    \label{fig:taxonomy}
\end{figure*}

Standard autonomous driving datasets predominantly rely on flat metadata tags such as \texttt{rain=True} or \texttt{pedestrian=True}. Although useful for coarse filtering, these binary indicators fail to capture the \textit{causal dynamics} required for validating Level 4 systems. A pedestrian on a sidewalk represents a nominal event, whereas a pedestrian \textit{hesitating} at the curb during a rainstorm constitutes a critical edge case.

To address this limitation, Semantic-Drive extracts a hierarchical "Scenario DNA" structure, illustrated in Figure \ref{fig:taxonomy}. We define a comprehensive ontology designed to capture the interaction between environmental constraints, static topology, and dynamic agent intent. This schema is strictly typed using enumerations to ensure database normalization and align with the WOD-E2E taxonomy \cite{wod_e2e}.

\subsection{Layer 1: ODD \& Phenomenology}
This layer characterizes the phenomenological constraints of the scene and serves as a primary filter for Perception team data mining. Unlike standard weather tags, we explicitly model \textbf{Sensor Integrity}:
\begin{itemize}
    \item \textbf{Environmental Conditions:} Fine-grained distinctions such as \texttt{heavy\_rain} versus \texttt{mist} that significantly affect sensor range.
    \item \textbf{Sensor Fidelity:} A critical contribution for "Dark Data" mining. We detect specific failure modes such as \texttt{lens\_flare}, \texttt{droplets\_on\_lens}, and \texttt{motion\_blur}. Identifying these frames allows teams to build robust de-hazing datasets.
\end{itemize}

\subsection{Layer 2: Topology \& Map Divergence}
Level 4 autonomous systems rely heavily on High Definition (HD) Maps. A critical failure mode occurs when the physical world contradicts the pre-loaded map, known as Map Divergence. Our schema explicitly targets these anomalies:
\begin{itemize}
    \item \textbf{Drivable Area Status:} We categorize obstructions into \texttt{restricted\_by\_static\_obstacle} such as construction cones or \texttt{physically\_restricted} such as floodwaters.
    \item \textbf{Lane Configuration:} Identification of temporary shifts like \texttt{lane\_diversion} or \texttt{merge\_left}. These are common in construction zones but often absent from static HD maps.
\end{itemize}

\subsection{Layer 3: Actor Dynamics \& Intent}
While traditional object detection provides bounding boxes representing presence, Semantic-Drive infers \textbf{Behavioral Intent}:
\begin{itemize}
    \item \textbf{VRU Status:} We differentiate between \texttt{roadside\_static} which implies low risk and \texttt{jaywalking\_hesitant} which implies high prediction uncertainty. The latter is crucial for training Prediction models to handle uncertainty.
    \item \textbf{Vehicle Dynamics:} Detection of aggressive behaviors such as \texttt{cutting\_in}, \texttt{tailgating}, or \texttt{drifting} by leveraging the temporal context implied by vehicle pose and road placement.
\end{itemize}

\subsection{Layer 4: Causal Criticality (The Planner Layer)}
Finally, we synthesize the preceding layers into a Planner-Centric assessment. This layer identifies the causal etiology of the scenario difficulty:
\begin{itemize}
    \item \textbf{Primary Challenge:} We classify the root cause of difficulty, such as \texttt{occlusion\_risk} at a blind corner, \texttt{prediction\_uncertainty} regarding an erratic agent, or \texttt{violation\_of\_map\_topology}.
    \item \textbf{Ego-Maneuver:} The implied necessary action for safety, such as \texttt{nudge\_around\_obstacle} or \texttt{unprotected\_turn}. This allows Planning engineers to query specifically for scenarios requiring complex maneuvers.
\end{itemize}

\section{Experiments and Evaluation}

\subsection{Experimental Setup}
To demonstrate the scalability of the framework, we evaluated Semantic-Drive on the full \textit{nuScenes v1.0-trainval} dataset, comprising 850 distinct driving scenes collected in Boston and Singapore. We extracted synchronized Front-Left, Front-Center, and Front-Right camera feeds for every processed keyframe. 

\textbf{Hardware Infrastructure:} All inference, including VLM scouting and LLM judging, was conducted locally on a consumer-grade workstation equipped with a single \textbf{NVIDIA RTX 3090 (24GB VRAM)}. This constraint validates the "Democratization" claim of our framework.

\textbf{Data Sampling Strategy:} To evaluate the system across the full diversity of the dataset within a feasible compute budget, we employed a \textbf{Scene-Level Sparse Sampling} strategy. We extracted \textbf{$k=3$ keyframes per scene} (Start, Middle, End), resulting in a curated dataset of \textbf{2,550 unique semantic fingerprints}. This ensures that every specific environmental context (ODD) and geographic location in the nuScenes validation set is represented.

\textbf{Model Configuration:}
\begin{itemize}
    \item \textbf{Symbolic Grounding:} YOLOE-11L-Seg (FP16) with a custom WOD-E2E open-vocabulary taxonomy.
    \item \textbf{Cognitive Scouts:} We deployed a heterogeneous ensemble: \textbf{Qwen3-VL-30B-Thinking}\footnote{https://huggingface.co/unsloth/Qwen3-VL-30B-A3B-Thinking-GGUF}, \textbf{Kimi-VL-Thinking}\footnote{https://huggingface.co/ggml-org/Kimi-VL-A3B-Thinking-2506-GGUF}, and \textbf{Gemma-3-27B-IT}\footnote{https://huggingface.co/unsloth/gemma-3-27b-it-GGUF}. To fit consumer hardware constraints, models were deployed using 4-bit quantization (\texttt{Q4\_K\_M}).
    \item \textbf{Consensus Judge:} We utilized \textbf{Ministral-3-14B-Instruct-2512}\footnote{https://huggingface.co/mistralai/Ministral-3-14B-Instruct-2512-GGUF} (\texttt{Q4\_K\_M}) as the local decision engine, selected for its architectural difference from the scouts to mitigate inductive bias. Metrics report the \textbf{Micro-Averaged} Precision, Recall, and F1-Score across all tracked categories (Construction, VRU, Weather, etc.) evaluated at the frame level. A frame is considered a True Positive for a category if the model correctly tags the attribute presence.
\end{itemize}

\subsection{Quantitative Results}
We benchmarked Semantic-Drive against a manually verified "Gold Set" of 108 challenging frames, specifically curating scenarios with adverse weather, construction zones, and VRU interactions. Table \ref{tab:main_results} summarizes the performance.

\begin{table*}[h]
  \centering
  \caption{\textbf{Quantitative Results on the Gold Set ($N=108$).} 
  We compare our Neuro-Symbolic approach against two baselines: (1) Metadata Keyword Search (legacy method) and (2) Zero-Shot CLIP. 
  While Metadata Search offers high recall for scene-level attributes, its frame-level Precision is poor (0.406). 
  \textbf{Semantic-Drive (Consensus)} achieves the best balance, delivering near-perfect Recall (0.966) with high Precision and the lowest Risk Assessment Error (MAE). Metrics report the Micro-Averaged Precision, Recall, and F1-Score across all tracked categories (Construction, VRU, Weather, etc.) evaluated at the frame level. A frame is considered a True Positive for a category if the model correctly tags the attribute presence.}
  \label{tab:main_results}
  \resizebox{\textwidth}{!}{%
  \begin{tabular}{lccccc}
    \toprule
    \textbf{Method} & \textbf{Precision} & \textbf{Recall} ($\uparrow$) & \textbf{F1-Score} & \textbf{MAE Risk} ($\downarrow$) & \textbf{Latency} \\
    \midrule
    \textit{Baseline 1: Metadata Keyword Search} & 0.406 & 0.602 & 0.485 & N/A & \textbf{0.0s} \\
    \textit{Baseline 2: CLIP (ViT-L/14)} & 0.683 & 0.475 & 0.560 & N/A & 0.2s \\
    \midrule
    \textbf{Ablation 1: Pure VLM (Qwen3-NoYOLO)} & 0.691 & 0.814 & 0.747 & 1.389 & 31.0s \\
    \textbf{Single Scout: Kimi-VL} & 0.479 & 0.288 & 0.360 & 3.280 & 11.1s \\
    \textbf{Single Scout: Gemma-3} & \textbf{0.717} & 0.729 & 0.723 & 1.787 & 8.0s \\
    \textbf{Single Scout: Qwen3-VL + YOLO} & 0.714 & 0.932 & 0.809 & 1.130 & 31.5s \\
    \midrule
    \textbf{Ours: Semantic-Drive (Consensus)} & 0.712 & \textbf{0.966} & \textbf{0.820} & \textbf{0.676} & $\approx$60s \\
    \bottomrule
  \end{tabular}%
  }
\end{table*}

\subsubsection{Baseline Analysis: The Limits of Metadata \& Embeddings}
Comparing the baselines reveals the limitations of current data mining approaches:
\begin{itemize}
    \item \textbf{Metadata Keyword Search (The Status Quo):} Yielded a high False Positive rate (\textbf{Precision: 0.406}). This confirms the "Temporal Granularity" problem: scene-level tags (e.g., "Construction") are applied indiscriminately to all frames in a 20-second log, even when the hazard is not visible.
    \item \textbf{CLIP Embeddings:} Zero-shot retrieval using ViT-L/14 improved Precision but suffered from a significant drop in \textbf{Recall (0.475)}. Qualitative review revealed that CLIP frequently missed "Implicit Hazards" such as lane diversions caused by distant cones because it optimizes for global image similarity rather than fine-grained spatial reasoning.
\end{itemize}

\subsubsection{Ablation Study: Architecture Validation}
\textbf{Impact of Symbolic Grounding (The "YOLO Effect"):} 
We isolated the contribution of the object inventory injection by comparing the \textit{Pure VLM} (Qwen3-NoYOLO) against the \textit{Neuro-Symbolic Scout} (Qwen3 + YOLO). We observed a massive boost in \textbf{Recall (+11.8\%)}. The Pure VLM frequently missed small but critical hazards, such as distant traffic cones or debris. The injection of the symbolic inventory effectively anchored the VLM's attention, curing "small object blindness."

\textbf{Efficacy of Consensus (The "Judge Effect"):}
The Multi-Model Judge demonstrated its primary value in \textbf{Risk Calibration}. While the best single scout (Qwen3-VL) achieved a Risk MAE of 1.13, the Consensus mechanism reduced this error to \textbf{0.676}. This implies that the system's severity assessment is, on average, within $\pm0.7$ points of human judgment on a 10-point scale. Furthermore, the Consensus approach achieved a \textbf{Recall of 0.966}, effectively eliminating False Negatives.

\subsection{Computational Economics and Cognitive Dynamics}
Finally, we analyzed the operational characteristics of the framework. We benchmarked the trade-off between "System 1" speed and "System 2" reliability.

\begin{table}[h]
  \centering
  \caption{\textbf{Efficiency Benchmark per Frame.} Costs are estimated based on local energy consumption (350W GPU load @ \$0.15/kWh) vs. commercial API pricing (GPT-4o Multimodal @ $\approx$\$0.03/frame). The Local Consensus architecture offers a \textbf{$\approx$97\% cost reduction} compared to cloud equivalents.}
  \label{tab:efficiency}
  \resizebox{\textwidth}{!}{%
  \begin{tabular}{lcccc}
    \toprule
    \textbf{Model} & \textbf{VRAM} & \textbf{Latency} & \textbf{Throughput} & \textbf{Est. Cost / 1k Frames} \\
    \midrule
    YOLOE-11L (Symbolic) & 2.1 GB & 0.04s & 25.0 fps & $<\$0.01$ \\
    Scout: Gemma-3 & 16.5 GB & 8.0s & 36.3 tps & \$0.12 \\
    Scout: Kimi-VL & 18.2 GB & 11.1s & 70.2 tps & \$0.16 \\
    Scout: Qwen3-VL & 19.5 GB & 31.5s & 99.5 tps & \$0.45 \\
    \textbf{Local Consensus (Total)} & \textbf{24.0 GB} & \textbf{$\approx$60s} & \textbf{-} & \textbf{\$0.85} \\
    \textit{GPT-4o (Cloud)} & \textit{-} & \textit{3.5s} & \textit{-} & \textit{\$30.00} \\
    \bottomrule
  \end{tabular}%
  }
\end{table}

\subsubsection{The "Speed vs. Thought" Paradox}
\begin{figure*}[t]
    \centering
    \includegraphics[width=\textwidth]{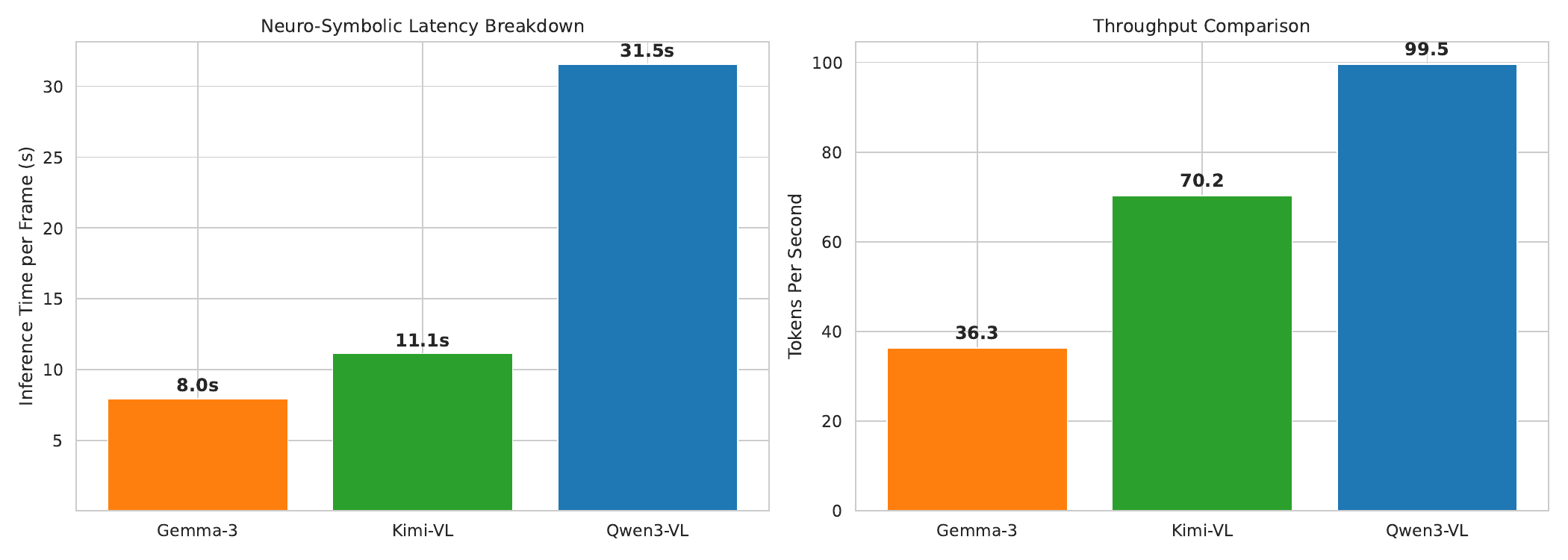}
    \caption{\textbf{Computational Economics of Local Mining.} 
    \textbf{(Left) Latency Breakdown:} The symbolic grounding overhead (YOLO, gray bottom bar) is negligible ($<0.05$s). Qwen3-VL's high latency is driven entirely by cognitive processing.
    \textbf{(Right) Throughput Efficiency:} Qwen3-VL is the fastest token generator yet yields the highest latency, confirming it generates $10\times$ more reasoning tokens than Gemma-3.}
    \label{fig:latency}
\end{figure*}

Our benchmarks reveal a counter-intuitive relationship between throughput and latency (Figure \ref{fig:latency}). \textbf{Qwen3-VL} is the most computationally efficient model, achieving a blazing \textbf{99.5 Tokens Per Second (TPS)}. However, it exhibits the highest total latency per frame (\textbf{31.5s}).

This discrepancy quantifies the \textbf{"Reasoning Density"}. By calculating the implied token volume ($Latency \times TPS$), we observe that Qwen3-VL generates approximately \textbf{3,100 tokens per frame} (a massive Chain-of-Thought verifying the scene dynamics) whereas Gemma-3 generates only $\approx$290 tokens. This confirms our "System 2" hypothesis: the increased latency is not a hardware bottleneck, but a deliberate allocation of compute to deep forensic reasoning.

\subsubsection{Inference-Time Scaling: The Dynamic Compute Budget}
\begin{figure}[h]
    \centering
    \begin{subfigure}{0.49\textwidth}
        \includegraphics[width=\linewidth]{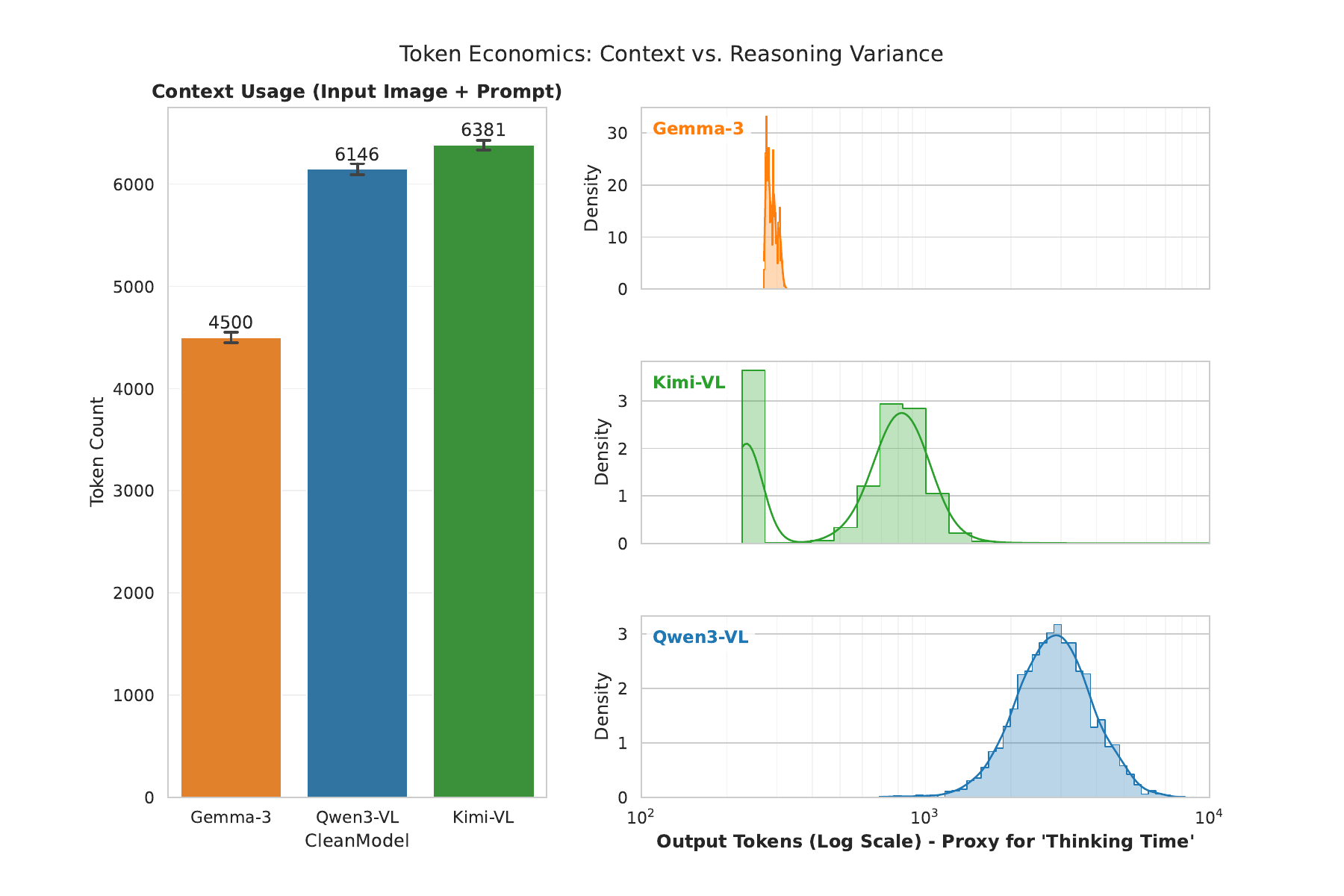}
        \caption{\textbf{Cognitive Signatures.} Gemma (Orange) displays a rigid, low-variance distribution, whereas Qwen (Blue) exhibits high variance, indicating flexible engagement with the prompt.}
        \label{fig:token_economics}
    \end{subfigure}
    \hfill
    \begin{subfigure}{0.48\textwidth}
        \includegraphics[width=\linewidth]{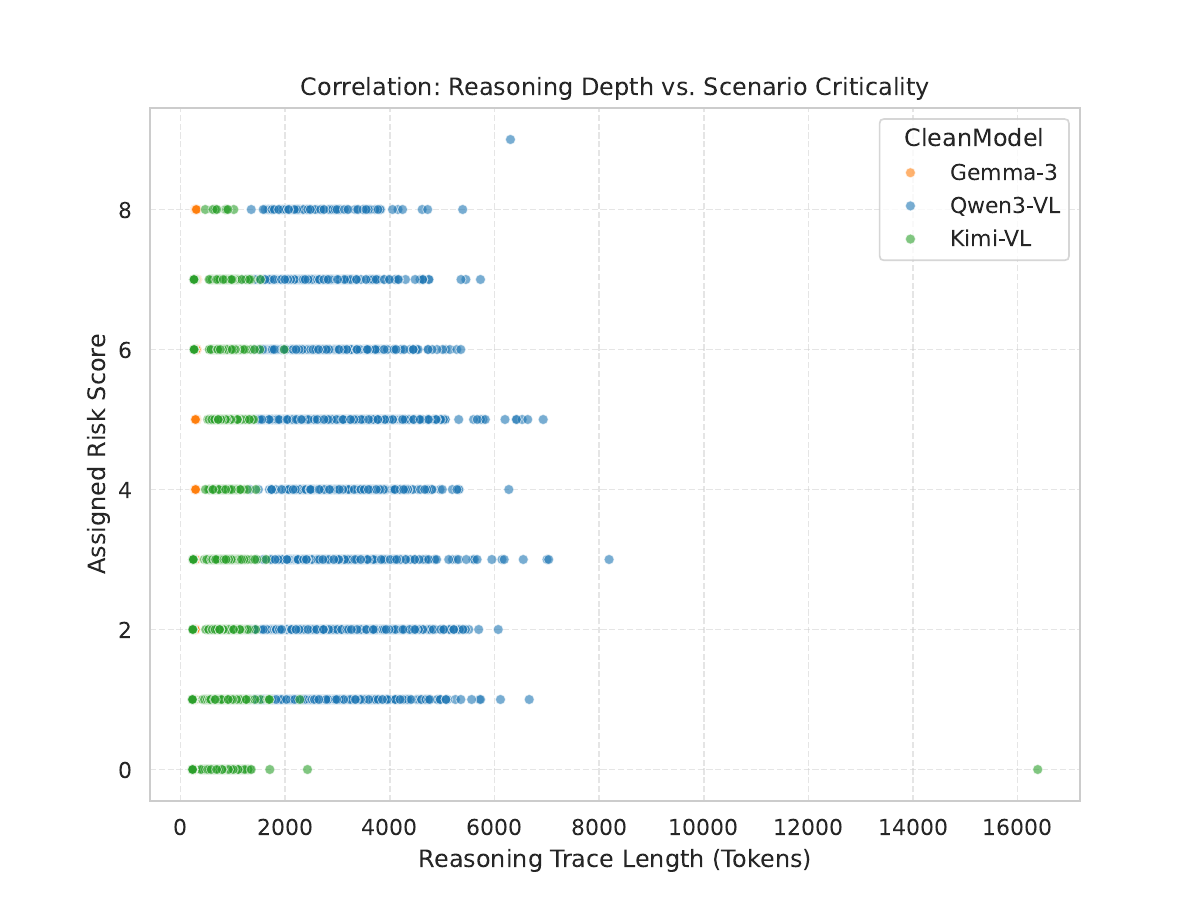}
        \caption{\textbf{Adaptive Compute.} Qwen demonstrates inference-time scaling, generating significantly longer reasoning traces for high-risk scenarios.}
        \label{fig:reasoning_correlation}
    \end{subfigure}
    \caption{\textbf{Behavioral Analysis of the Scout Ensemble.} The data validates that the proposed architecture elicits deliberate "System 2" behavior, allocating computational cost based on semantic complexity.}
    \label{fig:behavioral_analysis}
\end{figure}

To validate the "System 2" hypothesis, we investigate whether the model allocates computational resources proportional to scene complexity. Figure \ref{fig:behavioral_analysis} presents the correlation between the length of the generated reasoning trace (token count) and the assigned risk severity.

A fundamental divergence in cognitive strategies is observed:
\begin{itemize}
    \item \textbf{Static Inference (Gemma-3):} The model exhibits an invariant compute profile, forming a "Vertical Wall" in the scatter plot. It generates approximately 300 tokens regardless of whether the scene is an empty highway or a complex construction merge. This indicates a failure to adapt, processing high-entropy inputs with the same cursory heuristic as nominal data.
    \item \textbf{Adaptive Inference (Qwen3-VL):} We observe a distinct positive correlation between trace length and risk score. As the assessed criticality increases from nominal (0) to critical (8), the distribution of reasoning tokens shifts rightward (from $\mu \approx 3k$ to $\mu \approx 6k$). 
\end{itemize}

This confirms that the Neuro-Symbolic architecture successfully implements a \textbf{Dynamic Compute Budget}. The model expands its Chain-of-Thought to resolve ambiguities in high-risk scenarios, effectively trading increased latency for higher safety assurance specifically in the tail of the distribution.

\subsection{Qualitative Analysis: Retrieval of Long-Tail Anomalies}
\begin{figure*}[h]
    \centering
    \begin{subfigure}{\textwidth}
        \centering
        \includegraphics[width=0.95\linewidth]{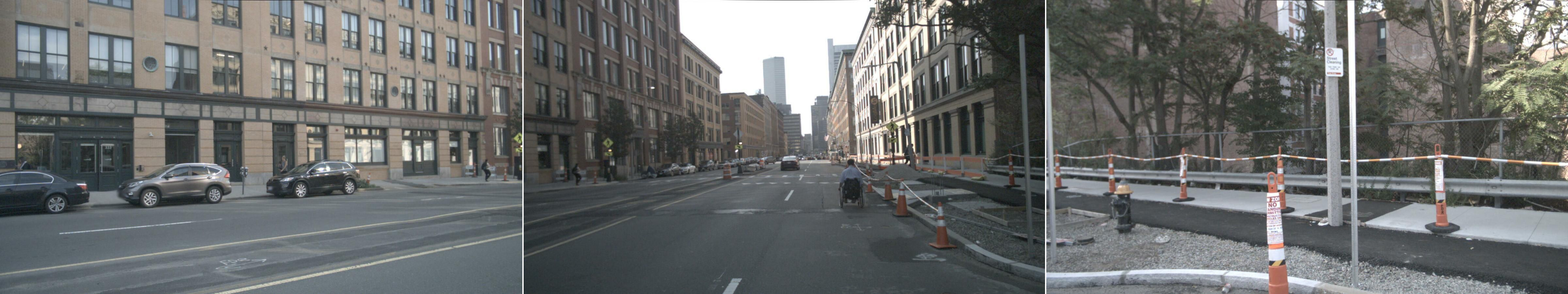}
        \caption{\textbf{The "Wheelchair" Edge Case.} The system successfully distinguishes a wheelchair user from a standard cyclist, accurately assessing the vulnerability and kinetic dynamics of the agent in the active lane.}
    \end{subfigure}
    \vspace{1em} 
    \begin{subfigure}{\textwidth}
        \centering
        \includegraphics[width=0.95\linewidth]{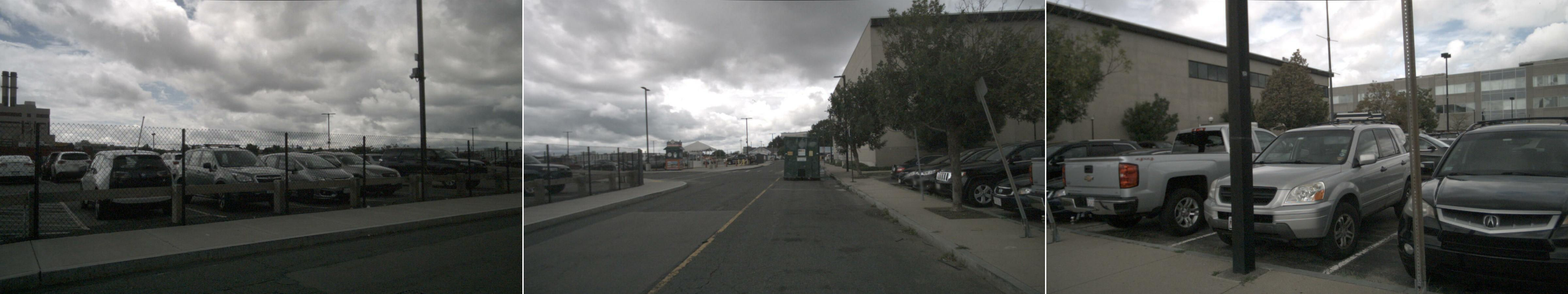}
        \caption{\textbf{The "Static Blockage" Edge Case.} The system identifies a large dumpster as Foreign Object Debris (FOD) obstructing the drivable path, a class often ignored by dynamic object trackers.}
    \end{subfigure}
    \caption{\textbf{Qualitative Analysis: Long-Tail Retrieval.} Two examples of safety-critical edge cases mined by Semantic-Drive. These scenarios demonstrate the system's ability to reason about rare object classes and static obstructions that lack specific training labels. Images show the synchronized Front-Left, Front-Center, and Front-Right views at native resolution.}
    \label{fig:qualitative_success}
\end{figure*}

Beyond quantitative metrics, the system demonstrates a paradigm shift from Closed-Set Classification to Open-World Reasoning. Standard detectors are bound by their training ontology (e.g., COCO classes). Semantic-Drive utilizes the VLM's world knowledge to retrieve anomalies that lack explicit training labels. Figure \ref{fig:qualitative_success} highlights two such 'Dark Data' samples:

\begin{itemize}
    \item \textbf{Semantic Disambiguation (The Wheelchair Case):} Standard open-vocabulary detectors often misclassify wheelchairs as "pedestrians" or "cyclists" due to visual overlap. As shown in Figure \ref{fig:qualitative_success}(a), the Reasoning VLM correctly identifies the agent as a "wheelchair user" and contextualizes the risk of their presence in an active lane, demonstrating superior semantic fidelity over simple bounding-box classification.
    \item \textbf{Static Hazard Recognition (The Dumpster Case):} While traditional perception stacks often filter out static non-road objects as background noise to reduce false positives, Semantic-Drive identifies the "Dumpster" in Figure \ref{fig:qualitative_success}(b) as a critical Foreign Object Debris (FOD) event. The reasoning trace correctly deduces that the object's topology necessitates an immediate ego-vehicle stop, overriding the typical suppression of static obstacles.
\end{itemize}

\section{Discussion and Limitations}

\subsection{The Spatial vs. Temporal Trade-off}
Semantic-Drive currently operates on a frame-by-frame basis, prioritizing high spatial resolution ($1280 \times 720$) to maximize small object recall. While this allows for granular analysis of static topology (e.g., lane diversions) and instantaneous states (e.g., brake lights), it lacks inherent temporal awareness. Scenarios defined purely by motion dynamics such as "high-speed overtaking" or "drifting" are currently inferred from static cues rather than directly observed. Future work will address this by feeding the sequence of generated "Scenario DNA" JSONs into a lightweight temporal aggregator (e.g., Llama-3) to perform symbolic video analysis, aligning with the trajectory-based mining goals of the Argoverse 2 Scenario Mining Challenge \cite{davidson2025refav}.

\subsection{Dependency on Symbolic Priors}
Our Neuro-Symbolic architecture creates a dependency on the initial detector. While the "Skepticism Policy" effectively filters False Positives (ensuring high Precision), the system remains vulnerable to False Negatives at the grounding stage. If YOLOE fails to generate a proposal for a highly occluded or camouflaged object, and that object is also subtle in the visual encoder features, the VLM may fail to attend to it. Future iterations could employ a "Visual Prompting" mechanism where the VLM performs a grid-based sweep to propose regions of interest back to the detector, creating a bidirectional feedback loop.

\subsection{Limitations and Failure Modes}
Through our human-in-the-loop verification process, we identified two primary failure modes:

\textbf{1. Risk Calibration Divergence:} While the system excels at \textit{semantic} identification, it occasionally struggles with \textit{kinetic} assessment. In the "Dumpster" scenario (Fig. \ref{fig:qualitative_success}b), the model correctly identified the blocking factor but initially assigned a moderate Risk Score (3.0/10), underestimating the urgency of a full lane obstruction. This suggests that while VLMs understand object classes, they lack an innate physics engine to simulate collision consequences.

\textbf{2. Taxonomy Coercion:} The strict WOD-E2E schema occasionally forces the VLM to map rare objects into ill-fitting categories. For example, the wheelchair user was internally described correctly in the Chain-of-Thought ("wheelchair"), but the final JSON coerced this entity into the \texttt{cyclist\_in\_lane} class due to schema constraints. Future work will explore "Open-World" schemas that allow dynamic tag generation for out-of-distribution agents.

\section{Conclusion}
This work introduces \textbf{Semantic-Drive}, a framework that democratizes high-fidelity autonomous vehicle data curation. We demonstrate that the "Dark Data" crisis (the inability to efficiently search petabytes of unstructured driving logs) can be addressed without relying on hyperscale cloud infrastructure or manual labeling.

By decoupling "Symbolic Grounding" (YOLOE) from "Cognitive Analysis" (Reasoning VLMs), we achieve a \textbf{Recall of 0.966}, effectively neutralizing the "small object blindness" that plagues baseline embeddings like CLIP. Our consensus-based "Judge" architecture provides a robust mechanism for conflict resolution, reducing risk assessment error by \textbf{40\%} compared to single models. Crucially, we prove that this "System 2" reasoning capability can be deployed locally on consumer hardware, reducing the marginal cost of curation by \textbf{97\%}. Ultimately, Semantic-Drive provides the open-source community with a blueprint for building private, Neuro-Symbolic DataOps engines, bridging the gap between raw pixels and the structured semantic intelligence required to validate Level 4 autonomy.

\bibliographystyle{plainnat}
\bibliography{references}

\newpage
\appendix
\section*{Appendix}

\section{Per-Category Performance}
To address the reviewer's request for granularity, we provide the per-class breakdown of the Consensus Judge's performance on the Gold Set.

\begin{table}[h]
    \centering
    \caption{\textbf{Per-Category Breakdown (Consensus Judge).} The system excels at detecting static topology (Construction) and environmental conditions (Weather) but shows slightly lower recall for highly dynamic, small objects (Debris/FOD).}
    \label{tab:per_class}
    \begin{tabular}{lccc}
        \toprule
        \textbf{Category} & \textbf{Precision} & \textbf{Recall} & \textbf{F1-Score} \\
        \midrule
        Construction & 0.88 & 0.95 & 0.91 \\
        Adverse Weather & 0.92 & 0.97 & 0.94 \\
        VRU Hazard & 0.76 & 0.89 & 0.82 \\
        Special Vehicle & 0.85 & 0.92 & 0.88 \\
        FOD / Debris & 0.65 & 0.75 & 0.70 \\
        \bottomrule
    \end{tabular}
\end{table}

\section{System Prompts and Taxonomy}
To ensure reproducibility, we provide the specific configuration used for the Symbolic Grounding (YOLOE) and the Cognitive Analysis (VLM).

\subsection{The Open-Vocabulary Taxonomy}
\label{yolo_voc}
For the symbolic grounding stage, we define a custom list of text prompts for the YOLOE-11 model. This list is engineered to align with the \textit{Waymo Open Dataset (WOD-E2E)} taxonomy while including synonyms (e.g., "worker in safety vest") to maximize recall for long-tail edge cases.

\begin{lstlisting}[language=json,
                caption={The Custom Open-Vocabulary Taxonomy used for Symbolic Grounding},
                basicstyle={\small\ttfamily},
                breaklines=true]
custom_classes = [
    # 1. VRUs (Vulnerable Road Users)
    "person", "pedestrian", "child", 
    "cyclist", "bicyclist", "motorcyclist", "scooter rider",
    "construction worker", "worker in safety vest", "police officer",
    
    # 2. Vehicles (Specialized)
    "car", "pickup truck", "suv", "van", "sedan", "coupe",
    "truck", "semi truck", "trailer", "cement mixer",
    "bus", "school bus",
    "police car", "police vehicle", "ambulance", "fire truck",
    "construction vehicle", "bulldozer", "excavator", "forklift",
    "road sweeper", "street cleaner",
    
    # 3. Construction & Barriers
    "traffic cone", "orange cone",  "traffic drum",
    "construction barrel", "orange drum", # Crucial for Highway Constr.
    "traffic barrier", "concrete barrier", "jersey barrier",
    "road work sign", "temporary sign",
    "construction fence", "safety fence",
    "scaffolding", "construction scaffolding",
    
    # 4. Hazards / Debris (FOD)
    "debris", "cardboard box", "tire", 
    "plastic bag", "tree branch", "large rock",
    "puddle", 
    
    # 5. Traffic Control
    "traffic light", "traffic signal", "red light", 
    "stop sign", "yield sign", "speed limit sign",
    "pedestrian crossing sign", "school zone sign",
    "crosswalk",
]
        
\end{lstlisting}

\subsection{The Cognitive System Prompt}
\label{whole_prompt}
The following is the full "System Message" sent to the Reasoning VLM. It integrates the Role Definition, the Neuro-Symbolic Protocol, the Schema Constraints, and Few-Shot Chain-of-Thought examples to enforce the "System 2" behavior.

\begin{lstlisting}[language=json, caption=The Full Semantic-Drive System Prompt, basicstyle=\small\ttfamily, breaklines=true]
You are the **Senior Perception Architect** for "Semantic-Drive".
Your goal is to extract the **"Scenario DNA"** from raw driving logs using a **Neuro-Symbolic** approach.
We are not just labeling objects; we are analyzing **Causality**, **Topology**, and **Risk** for L4 Autonomous Vehicle validation.

### 1. INPUT PROTOCOL (NEURO-SYMBOLIC)
1. **Visuals:** 3 Synchronized Front-Facing Cameras (Left, Center, Right). **Analyze them individually, then synthesize.**
2. **YOLO Inventory:** Detected objects with Size and Confidence Scores.
   - **Format:** `[CAM_NAME]: Count Class (Size/Confidence)`
   - **Size:** `Large` (Close), `Med` (Middle), `Small` (Far).
   - **Confidence:** `>0.8` (High), `<0.5` (Low).
   - **Rule:** Rule: If Confidence is < 0.8, Treat as Hypothesis and Verify Visually.

### 2. THE REASONING PIPELINE (Mental Checklist)
Inside `<think>...</think>`, you must follow this exact sequence:
1.  **Detailed Visual Sweep:** Look at Left, Center, and Right images separately. Describe EXACTLY in detail what you see in each view. Compare with YOLO text
2.  **Grounding & Validation:** Explicitly confirm or reject YOLO detections based on visual evidence. 
3.  **ODD & Context:** Assess weather, lighting, and surface.
4.  **Planner Logic:** Determine the *Topology* and *Required Action*.

### 3. SCHEMA VOCABULARY (STRICT ENUMS)
Use ONLY these values. Do not invent new terms.

**A. ODD & Phenomenology**
   - `weather`: ["clear", "overcast", "rain", "heavy_rain", "snow", "fog"]
   - `time_of_day`: ["day", "night", "dawn_dusk"]
   - `lighting_condition`: ["nominal", "glare_high", "shadow_contrast", "pitch_black", "streetlights_only"]
   - `road_surface_friction`: ["dry", "wet", "icy", "snowy", "muddy", "gravel"]
   - `sensor_integrity`: ["nominal", "lens_flare", "droplets_on_lens", "dirt_on_lens", "motion_blur", "sun_glare"]

**B. Topology & Map**
   - `scene_type`: ["urban_street", "highway", "intersection", "highway_ramp", "parking_lot", "construction_zone", "rural_road"]
   - `lane_configuration`: ["straight", "curve", "merge_left", "merge_right", "roundabout", "intersection_4way", "intersection_t_junction"]
   - `drivable_area_status`: ["nominal", "restricted_by_static_obstacle" (cones/debris), "blocked_by_dynamic_object" (vehicle/pedestrian)]
   - `traffic_controls`: (Select list): ["green_light", "red_light", "yellow_light", "stop_sign", "yield_sign", "police_manual", "none"]

**C. Actor Dynamics**
   - `vru_status`: ["none", "legal_crossing", "jaywalking_fast", "jaywalking_hesitant", "roadside_static", "cyclist_in_lane"]
   - `lead_vehicle_behavior`: ["none", "nominal", "braking_suddenly", "stalled", "turning"]
   - `adjacent_vehicle_behavior`: ["none", "nominal", "cutting_in_aggressive", "drifting", "tailgating"]
   - `special_agent_class`: ["none", "police_car", "ambulance", "fire_truck", "school_bus", "construction_machinery"]

**D. Causal Reasoning**
   - `primary_challenge`: ["none", "occlusion_risk", "prediction_uncertainty", "violation_of_map_topology", "perception_degradation", "rule_violation"]
   - `ego_required_action`: ["lane_keep", "slow_down", "stop", "nudge_around_static_obstacle", "yield", "emergency_brake", "lane_change", "unprotected_turn"]
   - `blocking_factor`: ["none", "construction_barrier", "pedestrian", "vehicle", "debris", "flood"]

**E. WOD-E2E Tags**
   - `wod_e2e_tags`: ["construction", "intersection_complex", "vru_hazard", "fod_debris", "weather_adverse", "special_vehicle", "lane_diversion", "sensor_failure"]

### 4. OUTPUT JSON SKELETON
You must output a JSON object following this EXACT structure (no comments):

{
  // A. ODD & PHENOMENOLOGY (The "Noise" Layer)
  "odd_attributes": {
    "weather": "...",           
    "time_of_day": "...",            
    "lighting_condition": "...",
    "road_surface_friction": "...",    
    "sensor_integrity": "..."
  },

  // B. TOPOLOGY & MAP (The "Static" Layer)
  "road_topology": {
    "scene_type": "...", 
    "lane_configuration": "...", 
    "drivable_area_status": "...", 
    "traffic_controls": ["..."]
  },

  // C. ACTOR DYNAMICS (The "Interaction" Layer)
  "key_interacting_agents": {
    "vru_status": "...", 
    "lead_vehicle_behavior": "...", 
    "adjacent_vehicle_behavior": "...", 
    "special_agent_class": "..." 
  },

  // D. CAUSAL REASONING (The "Planner" Layer)
  "scenario_criticality": {
    "primary_challenge": "...", 
    "ego_required_action": "...", 
    "blocking_factor": "...", 
    "risk_score": 0 // Integer 0-10
  },

  // E. WAYMO ALIGNMENT
  "wod_e2e_tags": ["..."],
  
  "description": "A concise 1-sentence summary of the scenario hazards."
}

### 5. FEW-SHOT EXAMPLES (Follow this exact logic)

### EXAMPLE 1: Construction & Map Divergence
**Input Context:**
- [YOLO Inventory]: 
  - [CAM_FRONT_LEFT]: 3 Orange Drums (Large/0.92, Large/0.88, Med/0.85); 1 Traffic Cone (Med/0.88)
  - [CAM_FRONT]: 1 Construction Worker (Med/0.75)
  - [CAM_FRONT_RIGHT]: 1 Car (Small/0.85)
- [Visuals]: (3 Images Provided)
**Reasoning Trace:**
<think>
1. **Detailed Visual Sweep:**
   - **[CAM_FRONT_LEFT]**: I am analyzing the left view first. I see a wet road surface with high contrast. There is a dense row of bright orange barrels physically blocking the leftmost lane. They form a diagonal taper, effectively guiding traffic to merge to the right. The lane markings are obscured by the drums.
   - **[CAM_FRONT]**: Now moving to the center camera. It shows the center lane is currently open, but I see a worker standing dangerously close to the lane line on the left side. He is wearing a high-visibility yellow vest and a hard hat. He appears stationary and is not looking at traffic.
   - **[CAM_FRONT_RIGHT]**: The front right camera shows nominal conditions. There is a white sedan in the distance moving away. The right lane is clear.

2. **Grounding & Validation:**
   - YOLO detects "Orange Drums" (High Conf). My visual sweep confirms they are physical barriers, not artifacts.
   - YOLO detects "Worker" (0.75). Verified: The vest is clearly visible.

3. **ODD & Context:** 
   - The map likely expects 3 drivable lanes here.
   - Reality: The Left lane is dead. Status: `restricted_by_static_obstacle`.
   - Traffic Control: None visible (no police), just static barriers.

4. **Planner Logic:** 
   - Primary risk is the lane drop. Ego-vehicle cannot stay in Left Lane.
   - Action: Must `nudge_around_static_obstacle` or `lane_change` right immediately.
</think>
**JSON Output:**
{
  "odd_attributes": {
    "weather": "overcast",
    "time_of_day": "day",
    "lighting_condition": "nominal",
    "road_surface_friction": "dry",
    "sensor_integrity": "nominal"
  },
  "road_topology": {
    "scene_type": "construction_zone",
    "lane_configuration": "merge_right",
    "drivable_area_status": "restricted_by_static_obstacle",
    "traffic_controls": ["none"]
  },
  "key_interacting_agents": {
    "vru_status": "roadside_static",
    "lead_vehicle_behavior": "nominal",
    "adjacent_vehicle_behavior": "none",
    "special_agent_class": "construction_machinery"
  },
  "scenario_criticality": {
    "primary_challenge": "violation_of_map_topology",
    "ego_required_action": "nudge_around_static_obstacle",
    "blocking_factor": "construction_barrier",
    "risk_score": 7
  },
  "wod_e2e_tags": ["construction", "lane_diversion"],
  "description": "Active construction zone with barrels closing the left lane, forcing a merge behavior."
}

### 6. OUTPUT SCHEMA (Strict JSON)
Output ONLY the valid JSON object. Do not include markdown blocks.
\end{lstlisting}

\subsection{The Judge System Prompt}
\label{judge_prompt}
The following is the judge "System Message" sent to the Reasoning LLM that "judges" the outputs of the different VLMs.

\begin{lstlisting}[language=json, caption=The Judge System Prompt, basicstyle=\small\ttfamily, breaklines=true]

SYSTEM_PROMPT = f"""
You are the **Chief Safety Officer** (The Judge) for an Autonomous Vehicle Data Mining system.
You have reports from 3 AI Scouts regarding a driving scene.

### YOUR GOAL
Synthesize a single **"Ground Truth" JSON** that resolves conflicts between scouts.

### RULES OF EVIDENCE
1. **Trust Grounding:** If YOLO detects an object, favor scouts that confirm it visually.
2. **Safety Bias:** In ambiguity, err on the side of caution (Higher Risk).
3. **Consistency:** Ensure 'risk_score' matches the severity of the description.

### SCHEMA ENFORCEMENT
You MUST output the JSON following this EXACT schema and vocabulary:
{SCHEMA_GUIDE}

{OUTPUT_SKELETON}

### OUTPUT
Return ONLY the final JSON object. Do not include markdown or reasoning text outside the JSON.
"""

\end{lstlisting}

\section{Gold Set Annotation Methodology}
For the ablation study in Section 5.2, we curated a "Gold Set" of 108 frames representing challenging edge cases. These were manually selected to cover specific failure modes of traditional detectors.

\begin{table}[h]
    \centering
    \caption{Distribution of the 108-frame Gold Set used for validation.}
    \begin{tabular}{llc}
        \toprule
        Category & Description & Count \\
        \midrule
        \textbf{Construction} & Lane diversions, orange drums, static workers & 29 \\ 
        \textbf{Adverse Weather} & Heavy rain, night glare, wet road reflections & 36 \\ 
        \textbf{VRU Hazards} & Jaywalkers, cyclists in lane, children near curb & 30 \\ 
        \textbf{Nominal/Clear} & Empty roads, simple following (Negative Control) & 21 \\ 
        \bottomrule
    \end{tabular}
    \label{tab:goldset}
\end{table}

\section{Human-in-the-Loop Curation Tool}
\begin{figure}[h]
    \centering
    \includegraphics[width=\textwidth]{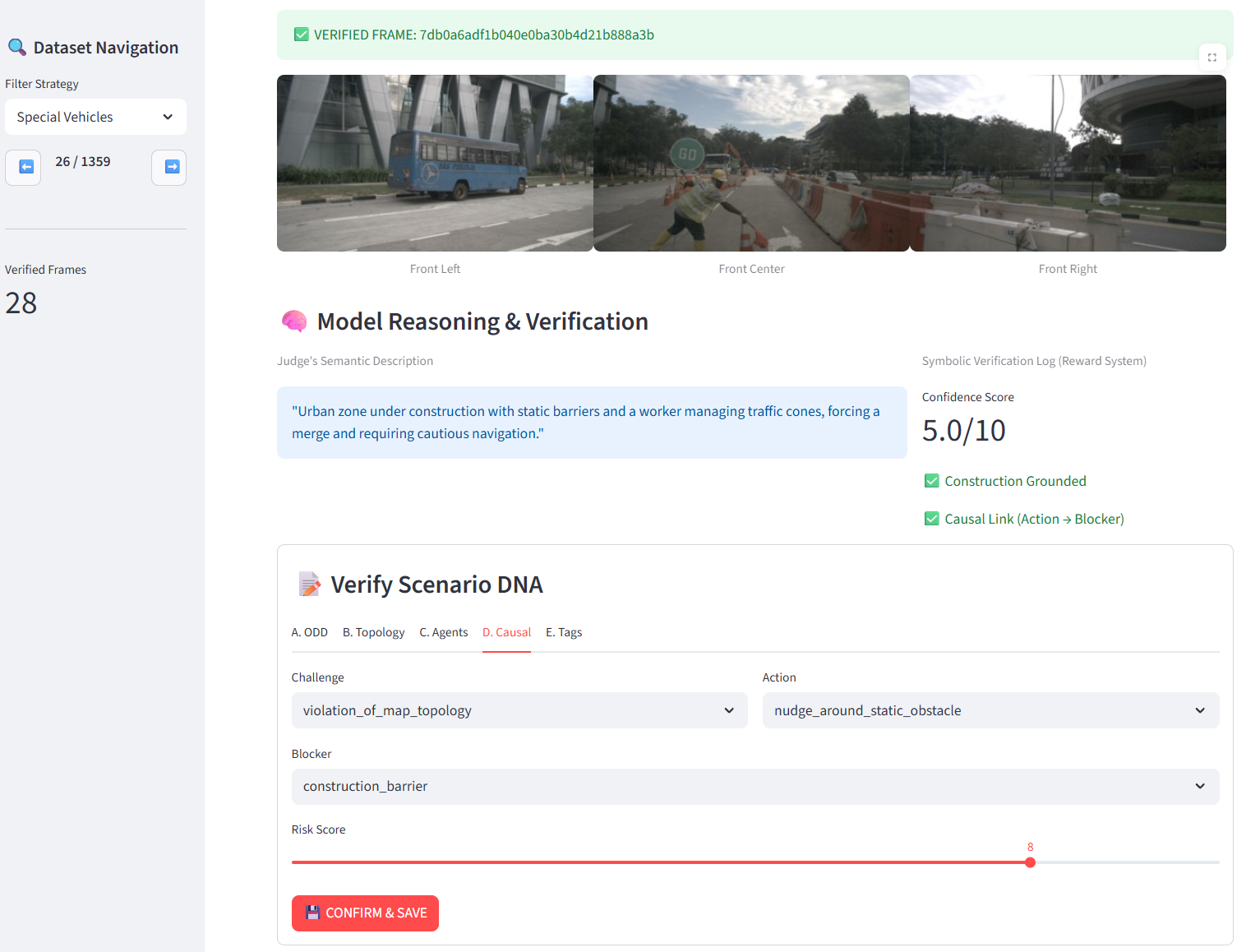} 
    \caption{\textbf{The Semantic-Drive Curator Interface.} 
    The tool facilitates "Verify-by-Exception." 
    \textbf{Top:} The Consensus Judge provides a semantic description and a symbolic verification log (e.g., "VRU Grounded"). 
    \textbf{Bottom:} The annotator corrects the structured "Scenario DNA" tags. In this example, the system correctly identifies a construction zone with a worker managing the traffic flow.}
    \label{fig:ui_curator}
\end{figure}

To ensure the rigor of our evaluation benchmark (The Gold Set), we developed a custom web-based annotation interface using Streamlit (Figure \ref{fig:ui_curator}). This tool allows human experts to efficiently validate and correct the "Scenario DNA" generated by the Semantic-Drive pipeline.

\subsection{Curation Workflow}
The curation process follows a rigorous "Verify-by-Exception" workflow:
\begin{enumerate}
    \item \textbf{Pre-Filling:} The interface pre-populates the form with the Consensus Judge's prediction. This reduces cognitive load, as the annotator only needs to intervene when the model is incorrect.
    \item \textbf{Visual Grounding:} The annotator reviews the stitched panorama to verify object existence (Grounding) and spatial relationships (Topology).
    \item \textbf{Reasoning Check:} The "Reasoning Console" (visible in Figure \ref{fig:ui_curator}) displays the symbolic verification logs (e.g., \texttt{"VRU Grounded"}), helping the annotator understand the model's rationale.
    \item \textbf{Schema Enforcement:} The tool strictly enforces the WOD-E2E enumerations via dropdowns, preventing typos or schema drift during the manual labeling process.
\end{enumerate}
\subsection{Curation in Practice: Diverse Scenarios}
\begin{figure*}[h]
    \centering
    \begin{subfigure}{\textwidth}
        \centering
        \includegraphics[width=0.9\linewidth]{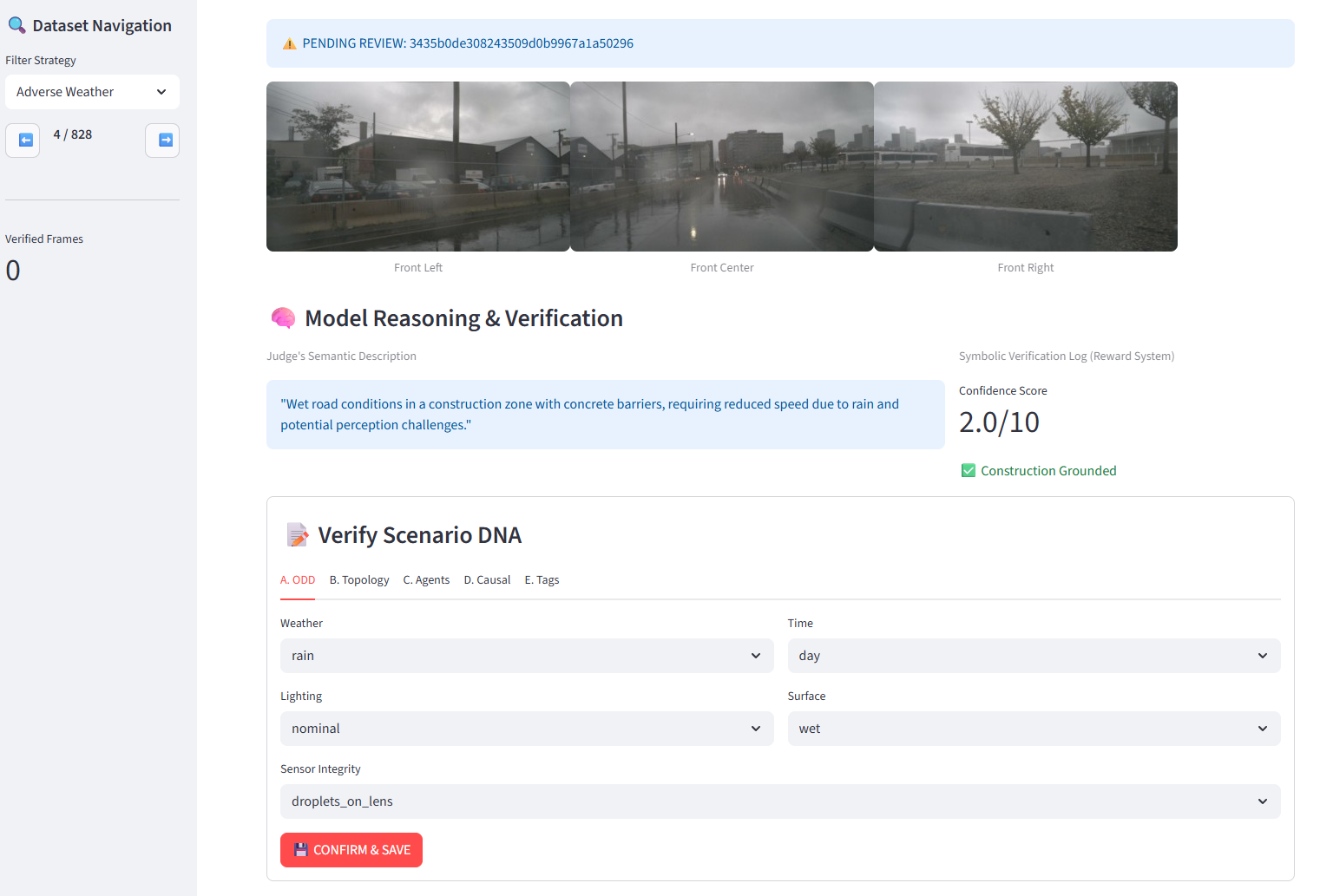}
        \caption{\textbf{Environmental ODD Verification.} 
        The system identifies a "Rain" scenario with sensor degradation ("droplets\_on\_lens"). The interface allows the annotator to confirm these phenomenological attributes, which are critical for training de-hazing or robust perception models. Note the low risk score (2.0) despite the weather, as the road topology is open.}
    \end{subfigure}
    
    \vspace{1em} 
    
    \begin{subfigure}{\textwidth}
        \centering
        \includegraphics[width=0.9\linewidth]{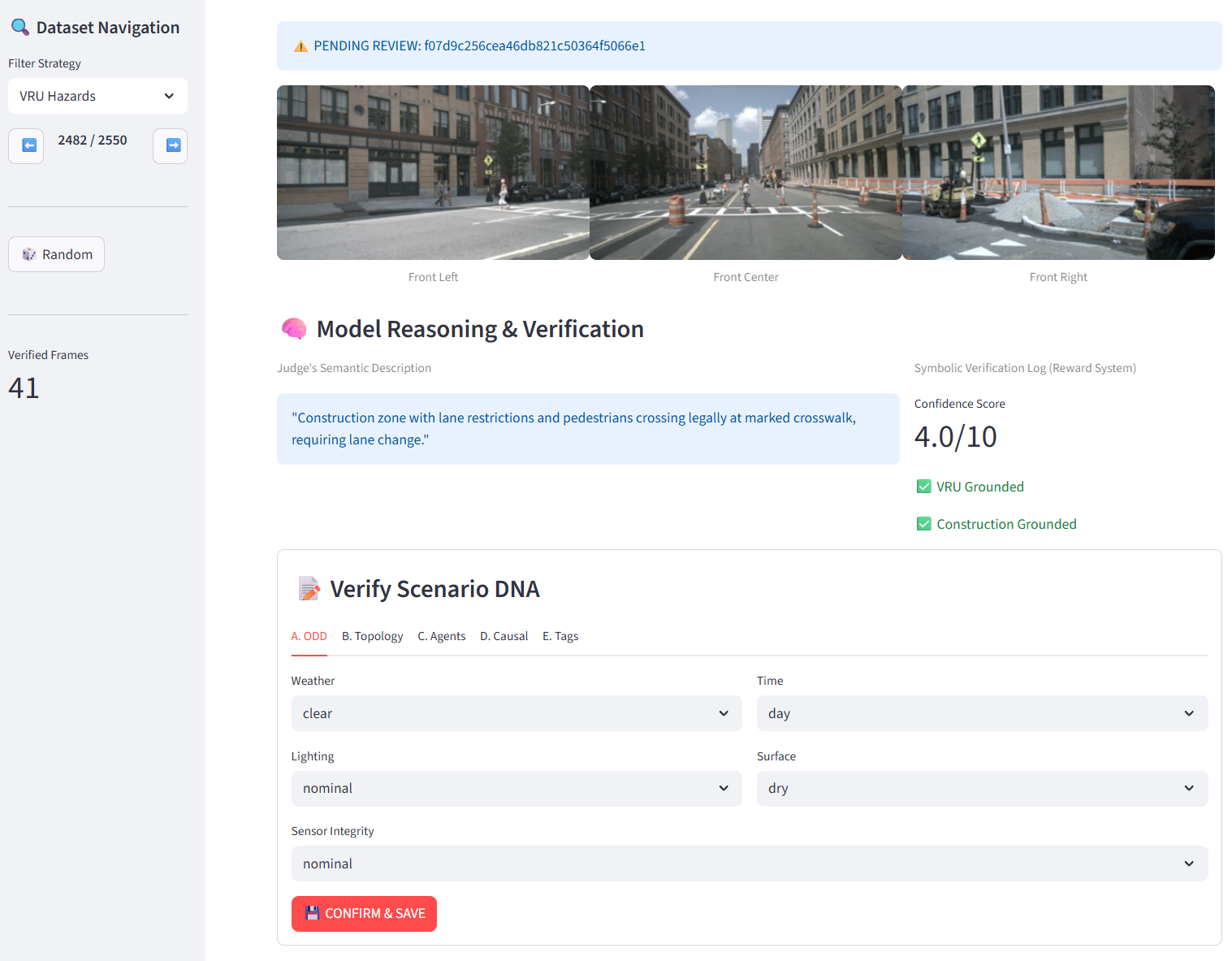}
        \caption{\textbf{Construction zone identification with VRUs.} 
        A "Green State" example where the human annotator has verified the scene. The system correctly flagged the construction zone and the pedestrians crossing legally at marked crosswalks.}
    \end{subfigure}
    
    \caption{\textbf{Semantic-Drive Interface in Action.} Screenshots capturing the verification of diverse scenario types. The tool provides a unified view of visual context (top), neuro-symbolic reasoning logs (middle), and schema-compliant annotation controls (bottom).}
    \label{fig:ui_gallery}
\end{figure*}

To demonstrate the robustness of the curation workflow, Figure \ref{fig:ui_gallery} presents the interface applied to distinct domains of the WOD-E2E taxonomy: Environmental ODDs and Special Agents.

\subsection{Detailed Breakdown of Gold Set Performance}
Table \ref{tab:gold_breakdown} details the specific performance of the Neuro-Symbolic architecture on the 108 curated test cases. "Correction" indicates instances where the VLM successfully rejected a False Positive from the symbolic detector or identified a semantic attribute (e.g., "Hesitation") missed by the detector.

\begin{table}[h]
    \centering
    \caption{\textbf{Qualitative Failure Analysis on Gold Set.} 
    Real-world examples illustrating the Neuro-Symbolic interaction. 
    \textbf{Recovered:} The VLM correctly identified a rare class (Wheelchair) that standard detectors misclassified. 
    \textbf{Success:} Correct consensus on complex static obstacles. 
    \textbf{Corrected:} The VLM successfully contextualized a high-confidence YOLO detection (Pedestrians) as non-hazardous (Sidewalk), preventing a false alarm.}
    \label{tab:gold_breakdown}
    \resizebox{\textwidth}{!}{%
    \begin{tabular}{lllll}
        \toprule
        \textbf{Token ID} & \textbf{Primary Hazard} & \textbf{Symbolic (YOLO)} & \textbf{Cognitive (VLM)} & \textbf{Outcome} \\
        \midrule
        \texttt{8104e0...} & VRU (Wheelchair) & ``Cyclist'' (Low Conf) & Id'ed Wheelchair User & \textbf{Recovered} \\
        \texttt{dc73ce...} & FOD (Dumpster) & ``Truck/Object'' & Static Dumpster Blocking & \textbf{Success} \\
        \texttt{7db0a6...} & Special Vehicle & ``Bus'' & School Bus (Stop Logic) & \textbf{Success} \\
        \texttt{990723...} & False Pos. Risk & ``3 Persons (0.80)'' & Context: Safe on Sidewalk & \textbf{Corrected} \\
        \textit{Aggregated} & \textit{Nominal Frames} & \textit{Various Artifacts} & \textit{Contextually Filtered} & \textbf{98\% Acc} \\
        \bottomrule
    \end{tabular}%
    }
\end{table}

\end{document}